\documentclass[twoside,11pt]{article}

\usepackage[preprint]{jmlr2e}

 \makeatletter
\renewcommand\normalsize{%
\@setfontsize\normalsize{11}{12.8}}
\makeatother
\normalsize

\usepackage[utf8]{inputenc} 
\usepackage[T1]{fontenc}    
\usepackage{hyperref}       
\usepackage{url}            
\usepackage{booktabs}       
\usepackage{amsfonts}       
\usepackage{nicefrac}       
\usepackage{microtype}      
\usepackage{xcolor}         

\usepackage{booktabs}       

\usepackage{pifont}   
\usepackage{marvosym}

\usepackage{xcolor,wrapfig}
\usepackage{hyperref}
\usepackage{amsmath,amssymb,amsfonts,graphicx}
\usepackage{latexsym}
\usepackage{textcomp}
\usepackage{graphics}
\usepackage{epstopdf}
\usepackage{multirow}
\usepackage{bm}

\usepackage{textcomp}
\usepackage{cleveref}
\usepackage{array}
\newcolumntype{?}{!{\vrule width 1pt}}
\usepackage{subcaption}

\usepackage{makecell}
\usepackage{algorithm}
\usepackage{algorithmic}

\def\be{\begin{equation}}
\def\ee{\end{equation}}

\jmlrheading{}{}{}{}{}{}{* These authors contributed equally}

\firstpageno{1}

\begin{document}
\setcitestyle{square,numbers}

\title{Symbolic Visual Reinforcement Learning: \\ A Scalable Framework with Object-Level Abstraction and Differentiable Expression Search}

\author{
\center
\name Wenqing~Zheng*,
        SP~Sharan*,
        Zhiwen~Fan, 
        Kevin~Wang,
        Yihan~Xi,
        Zhangyang~Wang \email \\ \{w.zheng, spsharan, zhiwenfan, kevinwang.1839, yihanx, atlaswang\}@utexas.edu \\
       \addr Department of Electrical and Computer and Engineering\\
       \addr The University of Texas at Austin, Austin, TX 78712, USA\\
       }

\editor{~}

\maketitle

\begin{abstract}
Learning efficient and interpretable policies has been a challenging task in reinforcement learning (RL), particularly in the visual RL setting with complex scenes. While neural networks have achieved competitive performance, the resulting policies are often over-parameterized black boxes that are difficult to interpret and deploy efficiently. More recent symbolic RL frameworks have shown that high-level domain-specific programming logic can be designed to handle both policy learning and symbolic planning. However, these approaches rely on coded primitives with little feature learning, and when applied to high-dimensional visual scenes, they can suffer from scalability issues and perform poorly when images have complex object interactions. To address these challenges, we propose \textit{Differentiable Symbolic Expression Search} (DiffSES), a novel symbolic learning approach that discovers discrete symbolic policies using partially differentiable optimization. By using object-level abstractions instead of raw pixel-level inputs, DiffSES is able to leverage the simplicity and scalability advantages of symbolic expressions, while also incorporating the strengths of neural networks for feature learning and optimization. Our experiments demonstrate that DiffSES is able to generate symbolic policies that are simpler and more and scalable than state-of-the-art symbolic RL methods, with a reduced amount of symbolic prior knowledge. Our codes are available at: \url{https://github.com/VITA-Group/DiffSES}.

\end{abstract}





\section{Introduction}\label{sec:introduction}

One major goal in reinforcement learning (RL) research is to develop generalizable, interpretable, and reliable policies \cite{kaiser2019model}. In recent years, deep neural networks (DNNs) have demonstrated great success in finding generalizable rules in various complex scenarios. However, the rules generated by DNNs are often criticized as being ``black boxes", which are difficult to interpret and trust \cite{mnih2013playing,heuillet2021explainability}.

To address the poor interpretability of DNN-based RL agents, symbolic reinforcement learning has emerged as a promising solution \cite{alibekov2016symbolic,kubalik2021symbolic,wilson2018evolving,junyent2019deep,bandres2018planning,lyu2019sdrl,ma2021learning,grounds2005combining,li2021interpretable,landajuela2021discovering,verma2018programmatically,garnelo2016towards,kimura-etal-2021-neuro,garcez2018towards,bastani2018verifiable}. Unlike DNNs, which parameterize the policy using learned neural activations and weights in a continuous high-dimensional space, symbolic RL composes the policy using a discrete combination of input operands and symbolic (mathematical) operators. For example, \cite{wilson2018evolving} proposed using Cartesian Genetic Programming to handle simple Atari games, and SDRL \cite{lyu2019sdrl} generated integrated symbolic planning to handle complex environments with high-dimensional inputs, such as Montezuma's Revenge. Compared to the policies generated by DNN agents and symbolic RL agents, symbolic policies are often more interpretable, lightweight, and efficient to execute.

Despite the successes of symbolic RL frameworks, scalability in complex visual scenes remains a challenge. This can be roughly attributed to two reasons as follows.

Firstly, more expressive image feature abstractions need to be explored. In the works that use symbolic expressions to directly generate controlling actions \cite{wilson2018evolving,miller2020cartesian,junyent2019deep,landajuela2021discovering}, the symbolic expressions are applied to raw input spaces. When the inputs are high-dimensional images instead of numerical states, the search for such rules becomes much more difficult or even intractable. This is because the input/output arities for general mathematical operators (e.g. "+, -, *, /") are small (with arities of two). To allow intelligent decision behaviors to emerge, it typically requires evolving into over-complicated architectures. For example, deep neural network agents pay the cost of complex layered matrix multiplications. Therefore, in order to scale to high-dimensional state spaces such as images, current works either define new operators with large arities (e.g. pure Cartesian Genetic Programming-based approaches \cite{wilson2018evolving,miller2020cartesian}), which limits the input to low-resolution images \cite{junyent2019deep}, or compose the existing operators into complex expressions \cite{landajuela2021discovering}. These approaches either reduce the interpretability of the resulting rules or leave issues unaddressed when scaling to truly high-resolution and complex image inputs. Furthermore, the high-dimensional space also introduces additional difficulty in optimization: as discussed in \cite{wilson2018evolving}, the learned symbolic policies with high-arity operators can be "disconcerting" and the optimization procedure is difficult to escape from local minima. To account for these issues and avoid composing overly complex symbolic policies, some symbolic RL frameworks heavily rely on pre-defined primitives. For example, \cite{wilson2018evolving} uses specifically designed matrix operators to process images, and \cite{lyu2019sdrl} plans according to human-defined high-level abstractions (e.g. what is a "ladder" and where is the "door"). These designs require significant human expert knowledge, so they may still lack the ability to automatically generalize to new environments with little human input.

Secondly, the poor scalability could aris from the lack of joint optimization mechanisms for architecture and coefficients. Being highly discrete architectures, symbolic expressions are difficult to learn through continuous optimization by nature. As a result, most symbolic RL frameworks have only established architecture learning, while feature learning may be insufficient \cite{landajuela2021discovering}, uninterpretable \cite{wilson2018evolving}, or unscalable \cite{junyent2019deep}. In non-visual RL tasks, existing symbolic RL frameworks usually generate end-to-end symbolic policies \cite{landajuela2021discovering}, where the symbolic architectures directly take the inputs of the states and output the final action, without preprocessing the input or any type of flexible feature learning. While these end-to-end approaches have been effective in non-visual tasks, in visual tasks, a certain level of feature learning and abstraction seems to be necessary to reduce redundant information in the images, extract key features, and provide lower-dimensional operands for the following symbolic architectures. In contrast, the way that current DNN-based RL agents handle feature learning is through layer-wise feed-forward and abstraction. In comparison, pioneering symbolic RL works have leveraged matrix operations \cite{wilson2018evolving} or cut the image into patches and summarized each patch separately \cite{coppens2019distilling}. To some extent, these hand-engineered, ad-hoc feature processing steps have reduced the interpretability of learned symbolic policies, their key advantage over DNN agents.


To address these challenges and improve the scalability of symbolic RL, we propose the Differentiable Symbolic Expression Search (DiffSES), which operates reliably under object-level representation (instead of pixel level) and jointly improves architecture and coefficient learning. The design of DiffSES is based on the assumption that object-level abstractions potentially yield simpler symbolic reasoning compared to raw pixel features \cite{samek2017explainable}. To this end, we set up feature learning as an unsupervised object detector that can summarize the image into a low-dimensional object feature space. After obtaining the object features, we use genetic programming (GP) to evolve the features into a symbolic expression that composes the action output based on these features and explainable mathematical operators. Unlike traditional GP, which generally suffers from low efficiency through brute-force search \cite{cranmer2020discovering,zheng2022symbolic}, we augment the optimization procedure in two ways: first, we make the coefficients trainable and implement gradient descent during evolution, and second, we propose a novel neural network-guided search procedure that alternates between optimizing a neural network and evolving symbolic trees. In the symbolic evolution process, we intentionally keep the mathematical operators as a small set to further reduce the dimensional object-level abstractions; the operand required primitives. Our contributions can be summarized as follows:

\begin{itemize}
\item We propose DiffSES, a novel symbolic RL framework that learns symbolic expressions as controlling policies based on object-level abstractions. The factorized object representations enable DiffSES to scale better to high-dimensional vision domains with less human expert knowledge.

\item To improve the generalization of complex vision domains, we propose a novel neural-guided symbolic evolution approach, which introduces gradient descent optimization into the symbolic expression fine-tuning and is the first to apply a neural-guided symbolic evolution for visual RL tasks.

\item Systematic comparisons with existing methods and extensive experimental results in Atari and Retro environments demonstrate the competitive performance of the proposed DiffSES framework.

\end{itemize}

\begin{small}

\begin{table*}[t]
\centering
\begin{tabular}[c]{p{0.2\textwidth}|p{0.5\textwidth}|p{0.15\textwidth}}
\toprule[1.5pt]
\textbf{Terminology} & \textbf{Short Definition} & \textbf{Reference Section} \\
\hline
Symbolic RL & The reinforcement learning approach that uses symbolic expressions to represent policies &  \Cref{sec:introduction} \\
\hline
Pixel-level representation & The raw matrix representation of images, holding height $\times$ width $\times$ channel pixels &  \Cref{sec:introduction} \\
\hline
Object-level representation & A summary of visual objects in a scene, rather than the raw pixel values & \Cref{sec:introduction} \\
\hline
Genetic programming (GP) & A type of evolutionary algorithm that evolves a population of candidate symbolic expressions through mutations/crossovers/etc., and finally pick the best one &  \Cref{sec:related works} \\
\hline
Operand & The math symbols that represent variables with values; appear as the leaf nodes in the symbolic tree & \Cref{sec:OD} \\
\hline
Operator & The math symbols used to connect other variables with values; appear as the branching nodes in the symbolic tree & \Cref{sec:OD} \\

\bottomrule[1.5pt]
\end{tabular}
\caption{The terminologies used in this paper.}
\label{tab:terminology}
\end{table*}
\end{small}

Terminologies used in this paper are shown in \Cref{tab:terminology}.

\section{Related Works}
\label{sec:related works}

The deep neural network-based RL agent, often referred to as DRL, parameterizes the policy via the learned neural network. Similarly, the symbolic-based RL agent, referred to as symbolic RL, or SDRL \cite{lyu2019sdrl}, parameterizes the policy via the symbolic expression. Besides the differences in the model composition, most other designed choices are shared across these two approaches: both are learned through the interaction with an environment, and the environments provide the same observations (states), and receive the action decided by either DRL or symbolic RL policies. This work intends to improve the flexibility of symbolic RL approaches and \textit{not} to compete with the DRL models.

The state-of-the-art symbolic RL algorithms could be categorized into the \textit{operator level} and \textit{reasoning level} methods. In the visual RL domain, the operator-level symbolic RL methods directly use symbols (math operators) to substitute the neural network parameterized policies in the DRL model \cite{wilson2018evolving,dittadi2020planning,junyent2019deep,bandres2018planning}, while the reasoning level symbolic RL uses symbols to represent higher-level abstractions, and guide the agent action selection or planning \cite{lyu2019sdrl,ma2021learning,grounds2005combining,li2021interpretable}. In the non-visual RL domain, most symbolic RL frameworks use operator-level symbolic policies to control vector-based state spaces effectively \cite{kubalik2021symbolic,alibekov2016symbolic,landajuela2021discovering,verma2018programmatically,garnelo2016towards,kimura-etal-2021-neuro,garcez2018towards,bastani2018verifiable}.

The operator-level symbolic RL are the ones that we mainly compare with. One major class of the operator-level symbolic RL methods adopts the Cartesian Genetic Programming (CGP) to evolve into a symbolic policy composed from a set of functions \cite{wilson2018evolving}. To allow for image processing, such functions include a class of matrix operations, for example, the skewness, kurtosis, mean, range, and other statistical or vector operations for the matrix of pixels. While being effective in controlling pixel inputs, it is hard to interpret the learned composition of CGP matrix operators. \cite{bandres2018planning} proposed to apply a width-based search algorithm on the B-PROST set of visual features. The algorithm, Rollout IW, was able to play Atari games comparable with humans. \cite{dittadi2020planning} improves on  Rollout IW and combines width-based planning with symbolic representation. The learning employs variational autoencoders (VAE) to learn relevant features in raw pixels of Atari games. Another recent work \cite{landajuela2021discovering} proposed a gradient-based approach that searches for symbolic representation. This further improves the model's efficiency and the discovered symbolic policies are readily interpretable.

The reasoning level symbolic RL methods improve the interpretation of reinforcement learning by indirectly integrating symbolic planning. For example, SDRL framework\cite{lyu2019sdrl} features a planner–controller–meta-controller architecture and lets the sub-layers in the controller learn based on intrinsic rewards. Another method, the NSRL\cite{ma2021learning}, for which the policy is induced to a neuro-logic reasoning module. Unlike the SDRL, this model extracts the logical rules instead of storing the rules. It is able to solve complex problems such as Montezuma's Revenge with the primitives of locations.

Existing visual symbolic RL methods can also be further divided based on where the learning occurs, which generally fall into two categories: \textit{search-based methods} and \textit{prediction-based methods}. In search-based methods \cite{wilson2018evolving,junyent2019deep}, the learnable component is the symbolic expression itself: through a carefully designed approach, stronger symbolic expressions are generated via smarter composition of symbolic operators/operands. This is typically done through a generation/selection procedure with genetic programming (GP) \cite{cranmer2020discovering} or other similar evolution methods. In prediction-based techniques, an independent learning agent learns to output the symbolic expression. This agent is usually a sequence prediction model such as a recurrent neural network \cite{landajuela2021discovering}. In this work, we adopt the \textit{search-based} symbolic generation method as a submodule of the proposed method.

In addition to the visual domain, other works have developed symbolic RL methods in vector-based state. One approach is to use symbolic regression to estimate the value function \cite{kubalik2021symbolic,alibekov2016symbolic} and derive a policy from the value function. This provides a mathematically tractable and interpretable policy, but the high computational complexity limits its application in higher-dimensional problems. \cite{verma2018programmatically} used a pretrained DRL model to direct a local search over symbolic policies that could generate human-readable policies. This approach requires manually setting specific prior syntactic models, but the resulting policies are able to learn smoother trajectories than neural policies.

\begin{figure*}[t]
    \centerline{\includegraphics[trim={0 3.5cm 0 3.5cm},clip,width=0.9\columnwidth]{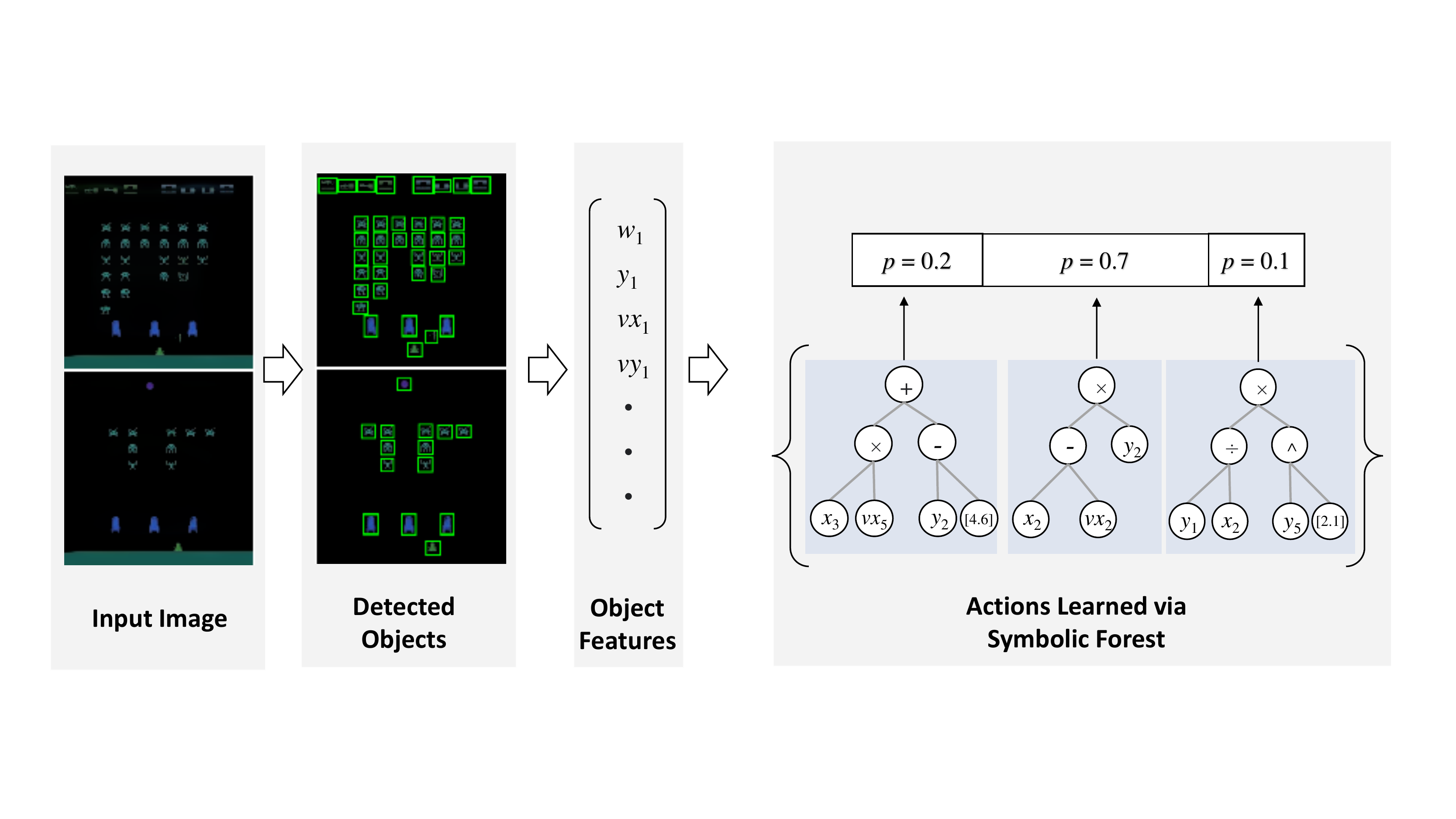}}
    \caption{The illustration of the forward inference procedure of DiffSES: the foreground objects are first captured through a self-supervised object detection module, and the object features are then processed by learned symbolic expressions.}
    \label{fig:flow}
    \end{figure*}

\section{Differentiable Symbolic Expression Search: Framework and Optimization}
\label{sec:approach}

\subsection{Preliminaries On Symbolic Expression And Symbolic Regression}
\label{sec:preliminaries}

\textbf{On The Tree/Forest Structured Symbolic Expressions.}
In general, a symbolic equation could be easily organized in the form of a tree \cite{petersen2019deep}, with the in-order traversal of this tree providing the equivalent string representation of the equation. A single symbolic tree is capable of accepting multi-dimensional \textit{inputs}, but its \textit{output} is usually restricted to a single scalar value. In order to scale the DiffSES symbolic tree to $\mathbb{R}^{N_A}$ dimensional action space, we learn $N_A$ such trees to compose a \textit{forest}. In this sense, each tree is responsible for one action, respectively. Similarly, it is straightforward to apply our method to $N_A$ dimensional continuous action space, or mixed continuous/discrete spaces, simply by letting all or some of the output dimensions represent those continuous actions.

As visualized in \Cref{fig:trees}, in each tree, the leaf nodes are the input object features or constants drawn from the \textit{operand space}, and the branch nodes are the math operators drawn from the \textit{operator space}. Each tree node possesses a float value. The value of a leaf node is the feature or constant value itself, and the value of a branch node is the execution result of the sub-tree, expanded from this branch node. Consequently, each tree also possesses a float value, which is the value of its root node. The resulting $N_A$ values of all trees represent the values of the corresponding actions. In the case of discrete action spaced environments, these values are the pre-softmax probabilities, following the common treatment of DRL models in the discrete action space as seen in \Cref{eq:action-continuous-discrete}, where $t_i$ is the value for $i$-th tree:

\begin{small}
\begin{equation}
    \texttt{action\ =}
    \begin{cases}
            \texttt{concat}_{i=1}^{A}[t_i], & \text{Cont. action env} \\
            \texttt{sample\ softmax}_{i=1}^{N_A}([t_i]), & \text{Disc. action env}
    \end{cases}
    \label{eq:action-continuous-discrete}
    \end{equation}
    \end{small}{}

With the symbolic forest formulation, we have a function that generates the output action probabilities based on the input observations. We define this function as the policy of this symbolic forest. The aim of our DiffSES agent is to learn this policy function to maximize the total reward.

\textbf{On Genetic Programming Based Symbolic Regression.}
As discussed in Section \ref{sec:related works}, we aim to develop a more flexible search-based symbolic RL approach that is learned at the operator level and directly controls the agent's actions. To achieve this, we will leverage the powerful and robust evolutionary algorithm, genetic programming (GP), as a submodule of our proposed method. GP maintains a group of candidate symbolic expressions and evolves them into new generations through variations such as random mutation and crossover. A screening process is used to select the best candidates, which become the next generation. This mutation/selection process is repeated until certain performance metrics are met or the maximum number of iterations is reached.

Naive symbolic search method apply this brute-force search procedure for one-time symbolic regression tasks without husstle. However, when applied to joint process of search and interaction, they may suffer from low efficiency if evolving from scratch for large-scale problems \cite{runarsson2000evolution,gustafson2005improving,orchard2016evolution,cranmer2020discovering,zheng2022symbolic}. It will become highly inefficient if further being entangled with the environment interaction, and may cause the searched results being biased to the initializations \cite{wilson2018evolving,real2020automl,runarsson2000evolution,orchard2016evolution}. Therefore, to accelerate symbolic search and avoid entanglement, new mechanism is needed to separate the policy learning and symbolic fitting subtasks.

\subsection{The DiffSES Framework}
\label{sec:framework}

Compared with DRL, symbolic RL is less flexible in exploration due to the additional burden of learning both architecture and coefficients. To address this issue and relieve the burden of exploration, exploitation, and fitting simultaneously, we propose a three-stage learning approach: \textit{neural policy learning}, \textit{symbolic fitting}, and \textit{fine-tuning}. In the first stage, we leverage the ease of continuous optimization of neural networks to learn a parameterized policy. In the second stage, we execute symbolic knowledge distillation to transfer the knowledge learned by the policy to a symbolic model (expression). In the final stage, we perform neural-guided search and fine-tuning to improve the performance of the symbolic model. Specifically, stage I will yield:

\be
\mathrm{Stage\ I:}\quad \bm{a} = f_\mathrm{neural}(\bm{x}; \theta)
\ee 
where $\bm{a}$ and $\bm{x}$ are the action and the image observation, $\theta$ is the neural network coefficients. The learned $f_\mathrm{neural}$ acts as the teacher model, and will be fixed once trained. In practice, we adopt the off-the-shelf PPO algorithm to train a standard CNN-based controller, while we also note that any neural network-based reinforcement learning algorithm could be used here. 

After training the neural net controller $f_\mathrm{neural}$, we proceed to the second stage of the symbolic fitting, where we discover the symbolic policy via symbolic regression module using GP. To achieve this goal, we need to define operand and operators. 

Previous symbolic RL approaches require extensive human expert knowledge to define high-level abstractions, such as labeling visual patterns, defining complex and domain-specific functional symbolic toolboxes, and assigning symbolic planning targets, etc. To learn the simple controlling symbolic expression, we follow the assumption: object level abstractions potentially yield simpler and more transferable symbolic reasoning compared with raw pixel features \cite{samek2017explainable}. Therefore, the symbolic expression takes the object level abstractions as its operands:

\be 
\label{eq:pipeline}
\bm{a} = f_\mathrm{symbolic}(\bm{x}'; \phi)
\ee 
where $\phi$ is the coefficient in the symbolic architecture, and the $\bm{x}'$ as the foreground object features. We will learn to detect these objects via self-supervised learning.

The learning procedure of $f_\mathrm{symbolic}$ follows a classical way of symbolic regression (SR), a type of regression analysis that searches the space of mathematical expressions to find an equation that best fits a dataset. Different from conventional regression techniques that optimize the parameters for a pre-specified model structure, SR infers both model structures and parameters from data. To run SR initially, one needs a dataset $\mathcal{D} = \textbf{X}, \textbf{Y}$, where $\textbf{X}\in\mathbb{R}^{N\times |\bm{x}'|}$ and $\textbf{Y}\in\mathbb{R}^{N\times {N_A}}$, where $N$ is the number of i.i.d. samples, $|\bm{x}'|$ is the total number of object features used, and $N_A$ is the dimension of action space. The symbolic regression procedure is then:

\be
\label{eq:sr}
\mathrm{Stage\ II:}\quad f_{\mathrm{symbolic}, i}, \phi_i = SR([\textbf{Y}]_{:,i},  f_{\mathrm{symbolic}, i}([\textbf{X}])
\ee

When the $f_\mathrm{symbolic}$ is fully learned, given a new input image $\bm{x}$, DiffSES first uses an object detection module to obtain foreground object features $\bm{x}'$, then pass these features into a forest of symbolic trees $f_\mathrm{symbolic}(\bm{x}')$ to decide the action. This procedure is visualized in \Cref{fig:flow}. 
After distilling the symbolic expression, we then use a fine-tuning stage to further optimize the coefficients and architectures of the learned symbolic expression, which we discuss in \Cref{sec:neural-guidance}.

\subsection{Details On Symbolic Fitting Stage II}
\label{sec:OD}

\textbf{On the object detection.}
In DiffSES, object-level features are extracted to serve as operands for symbolic expressions. To achieve this, we have several options for object detection (OD) algorithms, including template-based OD, supervised pretrained OD, or unsupervised OD. For greater flexibility in the subsequent symbolic learning, we adopt an unsupervised OD approach.

The object detection submodule in DiffSES is responsible for generating $\bm{x}'$ in \Cref{eq:pipeline}. We use Spatially Parallel Attention and Component Extraction (SPACE) \cite{lin2020space} to pre-train the foreground object detection submodule of DiffSES, which is then fixed. SPACE is an unsupervised object detection algorithm that unifies spatial attention and scene-mixture approaches without the need for manual labels. It consists of two streams: a foreground module responsible for detecting dynamic objects such as the main agent and entities, and a background module responsible for detecting the relatively static background of the environment. These definitions are learned implicitly by SPACE during training. The loss functions for the two streams are designed to optimize for "movement" across frames, under the assumption that game entities are relatively non-static compared to the static background. Given raw observations of the environment during training, the foreground and background streams decompose them into factorized representations of independent objects and segments, respectively. The distributions of these components are then combined using a pixel-wise mixture model to produce the complete image distribution. SPACE also overcomes scaling issues by using parallel spatial attention, making it suitable for scenes with a large number of objects. As a result, it combines the benefits of both scene-mixture and spatial-attention models.

As an interesting discovery in the experiments, we found that the unsupervised OD could not always yield satisfactory detection results from a human interpretation perspective, as it could, at times, separate a single object into several disjoint components. However, this non-satisfactory detection did not lead to DiffSES algorithm failure: the succeeding symbolic learning module ultimately only pick one of the splitted objects, and composes a robust symbolic expression with the previous OD as a whole. More discussions on the object detector submodule are in \Cref{sec:ablation}.

\textbf{On the operand selection.}
Before conducting symbolic regression, one need to define the corresponding operand space and operator space. The operand space is constructed from the features of the objects, which are provided by the unsupervised object detection module. This module generates object class, location, bounding box size, and moving velocity that are ready for use as operands for symbolic regression. Specifically, for each detected object, we append four features to the operand set: the $x$ coordinate, the $y$ coordinate, and the horizontal and vertical components of the velocity ($v_x$ and $v_y$). This means that $M$ detected objects will result in $4M$ features. In consecutive frames, the number of objects may remain the same, but new objects may appear from the edge of the frame or emerge from the center, resulting in a change in the number of objects. Regardless of the number of detected objects (which is usually large), we always filter the top $\bar{M}$ objects with the highest detection probability, with their type aligned across frames. In the rare cases where fewer than $M$ objects are detected, the operand features are padded with zeros.

\textbf{On the operator selection.}
The choice of object-level abstraction (instead of pixel-level representation) enables the use of simpler symbolic expressions to generate complex behaviors. This can be explained by the representation power of object representations. For example, when paired with some learned constants, simple operators such as $+$, $-$, $*$, and $/$ are sufficient to represent complicated spatial-temporal relationships between objects, such as "object A is on the colliding path of object B" or "object A is right above object B." Such relationships are often sufficient to control environments in Gym/Arcade, particularly shooting games or object avoidance tasks. To this end, we eliminate the use of higher-order statistical operators such as skewness and matrix variance as in \cite{wilson2018evolving}. We set the operator space to contain the following operators: $+$, $-$, $\times$, $/$, $\leq$, $\geq$, $\neg$, $(\cdot)^2$, $(\cdot)^3$, $\sqrt{\cdot}$, $\exp$, $\log$. We intentionally choose this operator space to be small and simple, which also reduces the required human expert knowledge when scaling to new environments.

\subsection{Multi-Action Optimization through Neural Guided Symbolic Search}
\label{sec:neural-guidance}


\begin{figure*}[t]
    \centerline{\includegraphics[trim={0 1.5cm 0 1.5cm},clip,width=0.9\columnwidth]{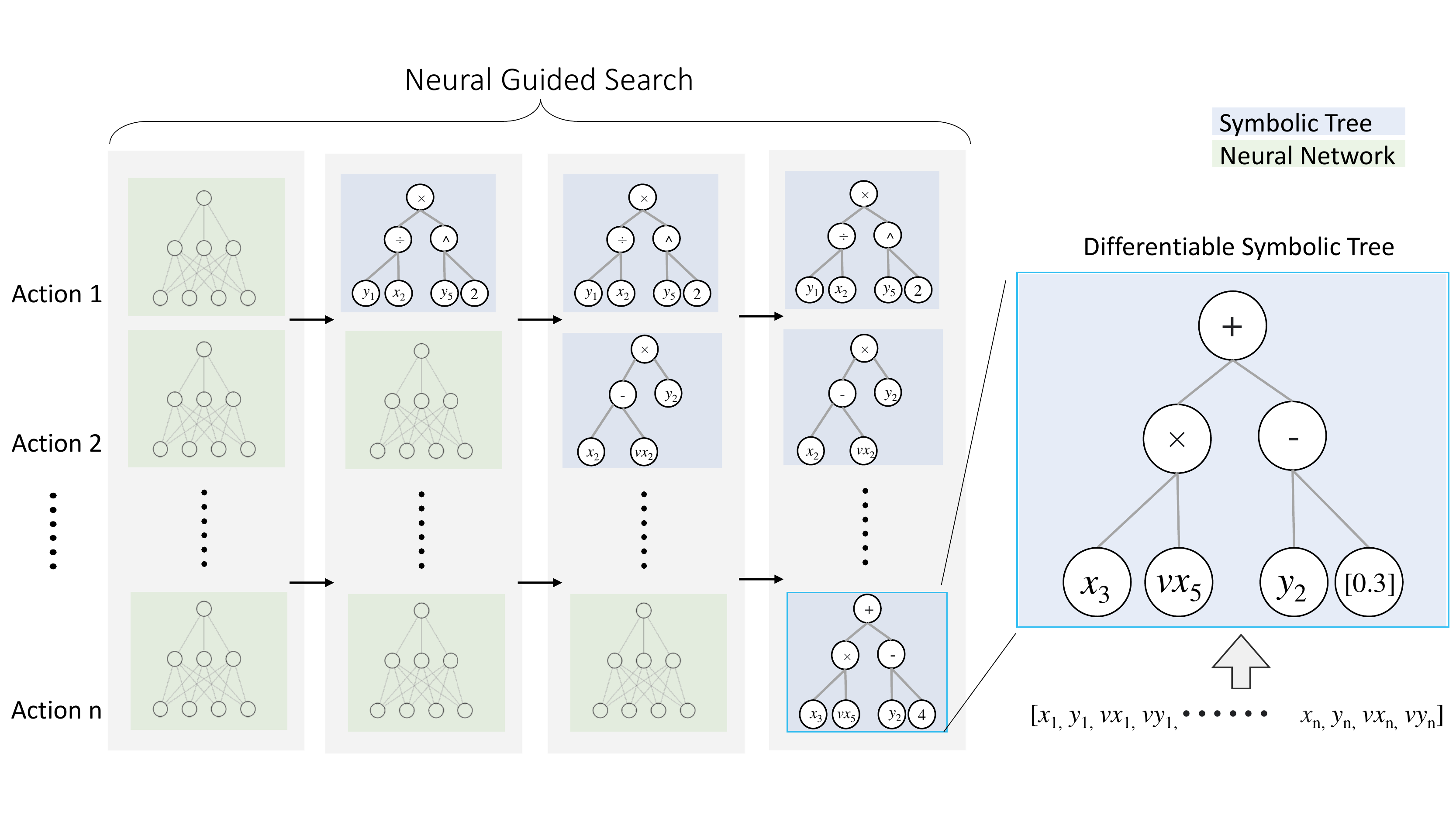}}
    \caption{
        The neural-guided search procedure of DiffSES. Initially, the agent's action is determined purely by the neural network. It then gradually learns the symbolic representation action by action.
    }
    \label{fig:trees}
    \end{figure*}

The neural guided multi-action symbolic policy search is a \textit{staged optimization} technique, it guides the symbolic evolution with pre-trained neural network models to evolve $N_A$ trees in the $N_A$ dimensional action space.

The traditional evolutionary algorithm such as genetic programming maintains a population of offsprings, and applies a mixture of mutation, crossover, pruning, and a few other operations to evolve the symbolic trees. In order to efficiently learn the symbolic forest, we propose a more flexible evolution method as the third stage of DiffSES. Compared with previous mutation based symbolic RL, the third stage of DiffSES presents two novel mechanisms: the \textit{gradient descent optimization} for the coefficients, and the \textit{neural guided multi-action tree search}, as visualized in \Cref{fig:trees}.

Unlike most neural networks which have fixed structures and one only needs to optimize their coefficients, the symbolic expressions need both skeletons (the structure of the expression) composition and coefficients (the ``constants'' in the expression) optimization. While the skeleton is hard to continuously optimize, the coefficients could receive gradients and be optimized continuously. We supplement the conventional genetic mutation by applying gradient descent for the coefficient of the symbolic expression. As the symbolic policy is just composed of mathematical operators, it is straightforward to obtain the computational graph for an output with respect to its scalar parameters for any given input. In other words, the evolution of the genetic program can now be supported by using the information on the derivatives, hence enabling the equivalent of back-propagation in Neural Networks.

Our proposed partially differentiable symbolic tree can be used as a drop-in replacement for a neural network in any DRL method. We adopt the PPO algorithm (explained in \Cref{alg:PPO}) as the backend of differentiable optimization, and apply SGD with 0.005 learning rate when it requires gradient descent. The differentiability feature is implemented as an additional evolution strategy alongside the existing evolution options (crossover, subtree mutation, hoist mutation, point mutation, and reproduction). We use 0.2 probability for the SGD, and keep the default ratios of the original GP evolution options \cite{stephens2019gplearn}, which share the rest 0.8 probability during the evolution tournament.

During stage III, the symbolic expressions starts to evolve from the symbolic regression results in stage II as warm initializations. During learning, only one tree is mutated at a time, while the other actions are controlled by the pretrained DRL agent (its teacher agent from stage I). We iteratively freeze the symbolic trees to start training another symbolic tree for the next action, replacing the corresponding DRL controller. \Cref{fig:trees} graphically depicts our proposed neural-guided search approach. In this way, the symbolic evolution procedure is eased by a more flexible neural network to improve the convergence rate of the symbolic expression.

The target function to be maximized is the same as the PPO algorithm, which takes the policy parameters $\theta, \phi$ (coefficients of DRL and symbolic RL) as inputs: 
\begin{small}
\begin{align}
\mathcal{L}(\theta, \phi) = \frac{1}{N} \sum_{i=1}^N \min(r_t(\theta, \phi) \cdot A_t, clip(r_t(\theta, \phi), 1-\epsilon, 1+\epsilon) \cdot A_t)
\label{eq:ppo}
\end{align}
\end{small}

Here, $r_t(\theta)$ is the ratio of the new policy to the old policy for a given experience, and $A_t$ is the estimated advantage for the experience:
\begin{small}
\begin{align}
A_t = -V(s_t)+r_{t} + \gamma r_{t+1} + \cdots+\gamma^{T-1+1} r_{T-1} + \gamma^{T-t}V(s_T)
\end{align}
\end{small}
where $\gamma$ is the discount factor determining the relative importance of future rewards.
The values are clipped to $1-\epsilon$ and $1+\epsilon$, where $\epsilon$ is a small positive value that controls the amount of clipping used in PPO. The overall effect of the loss function is to encourage the policy network (i.e., neural network with one action output replaced by symbolic expression) to select actions that maximize the expected reward while staying within a certain range of the old policy. This helps prevent the policy from making sudden, large changes that could destabilize the learning process. We follow these default settings from PPO without modifications. The algorithm is shown in \Cref{alg:PPO} which inherits from the PPO algorithm \cite{ppo}.

\begin{algorithm}
\caption{Neural-Guided Symbolic Forest Fine Tuning}
\begin{minipage}{0.8\textwidth}
\centering
\begin{algorithmic}[1]

\REQUIRE Pre-trained PPO teacher agent $\pi_{\theta}$, initialized symbolic expression forest $\pi_{\phi}$.
\ENSURE Optimized symbolic policy $\pi_{\phi}^*$
\STATE Initialize empty experience replay buffer $D$
\FOR{each iteration}
  \FOR{$i$-th action}
    \STATE Replace $i$-th output of ${[\pi_{\theta}]}_{i}$ with expression ${[\pi_{\phi}]}_{i}$
    \STATE Run mixed policy $\pi_{[\theta,\phi]}$ to generate $D$
    \STATE Sample a mini-batch of $(s_t, a_t, r_t, s_{t+1})$ from $D$
    \STATE Compute advantage $A_t$ for each experience
  \ENDFOR
  \STATE Update policy by maximizing the PPO objective \Cref{eq:ppo}
  \STATE Update old policy parameters: $[\theta,\phi]_{old} \leftarrow [\theta,\phi]$
\ENDFOR
\STATE Return the best performing symbolic expressions as the final policy $\pi_{\phi}^*$
\end{algorithmic}
\label{alg:PPO}
\end{minipage}
\end{algorithm}

\begin{figure*}
    \centering
    \includegraphics[width=0.9\textwidth]{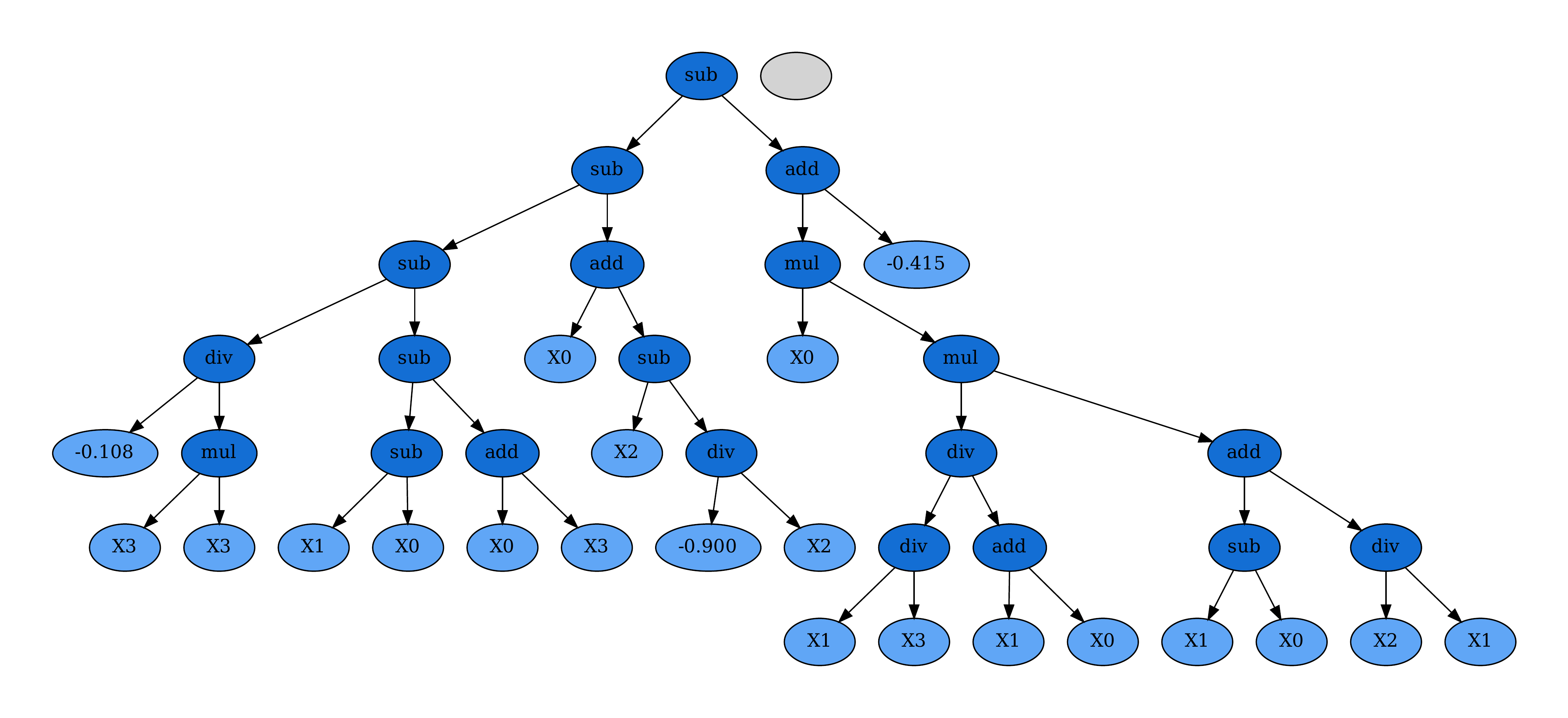}
    \vspace{-2em}
    \caption{Visualization of a trained DiffSES policy for the Pong-Atari2600 environment. 
The leaf nodes $X_{0}$, $X_{1}$, $\dots$ are the relabeled objects' positions and velocities. Such symbolic expressions offer potential explanability of the controlling policy: some subtrees might happen to constitute geometrically interpretable meanings.
For example, the geometric features (leaf nodes) $X_{0}$, $X_{1}$, $\dots$ could be  $x_\mathrm{pong}$, $x_\mathrm{racket}$, $y_\mathrm{pong}$, $v_\mathrm{y,pong}$, the x/y location of the pong and the racket, and vertical velocity of the pong. Then one subtree might appear as: $y_\mathrm{pong} + v_\mathrm{y, pong} * (x_\mathrm{racket} - x_\mathrm{pong})*c$. This could mean the $y$ axis of the aiming point of the pong on the racket, where $c$ is some constant to convert the horizontal distance into time gap. The aiming point is where the racket should ultimately go to. If such sub-expressions are found, it could hint that similar logic is learned.
    }
    \label{fig:policy-pong}
\end{figure*}

\section{Experimental Settings And Results}
\label{sec:exp}

In this section, we perform a systematic study of the DiffSES. First, we show hyperparameters settings and an example of the learned symbolic policies in \Cref{sec:viz-sub1}. Then, we perform systematic ablation studies for object detection module and neural guided search components in \Cref{sec:ablation}. Next, comparisons with existing symbolic RL methods are presented in \Cref{sec:exp:baseline}, in both visual and non-visual environments in Atari and Retro. Finally, a brief comparison between the learned symbolic policy and its neural network teacher is presented in \Cref{sec:compare-with-drl-human}.

\subsection{Experimental Settings And Visualizations}
    \label{sec:viz-sub1}
    When training the teacher model in stage I, we took the standard implementation from stable-baselines3 \cite{raffin2019stable} with reward decay $\gamma=0.9$, learning rate $0.0005$ to train a 6-layer CNN-based PPO agent. When distilling the symbolic policy in stage II, we use the gplearn \cite{stephens2019gplearn}. We select the features of the top $\bar{M} = 16$ objects with the highest probability, leading to 64 features. The number of samples is $N=2000$, the population as 50 for each iteration, and the total number of iterations as 300. When fine-tuning the symbolic expression with neural guided search in stage III, we adopt the same training setting as stage I, and the same symbolic hyperparameter as stage II, and loop for 50 iterations. Due to engineering facilitation considerations, we re-normalize the rewards to set the games to only have one life.

    The visualization of one learned policy tree resultant from the Pong-Atari2600 environment is shown in \Cref{fig:policy-pong}. More examples of symbolic trees are provided in \Cref{sec:appendix}. From performance side, the reward for the symbolic policy learned in four visual environments are shown in \Cref{tab:od-failure}. The teacher DRL model performance is also listed as a reference, yet we aim not to compare with it, as discussed in \Cref{sec:related works}.

    \begin{table}[h]
        \centering
        \begin{tabular}{c|c|c}
            \toprule[1.5pt]
            Environment & DiffSES Reward & DRL Reward \\
            \midrule
            CircusCharlie-Nes & $3549$ & $4580$\\
            AstroRoboSasa-Nes & $1256$ & $966$\\
            Seaquest-Atari2000 & $191$ & $208$\\
            Airstriker-Genesis & $421$ & $383$\\
            \bottomrule[1.5pt]
        \end{tabular}
        \caption{Results of DiffSES and DRL agents on four visual RL environments.}
        \label{tab:additional-visual-rl}
    \vspace{-1em}
  \end{table}

\subsection{Ablation Study}
\label{sec:ablation}

  \subsubsection{Ablation Study On Sub-optimal Unsupervised Object Detection Module}
    
    As the generated symbolic policy is reliant on the success of the object detection submodule (it require the detected objects as its input operands), one natural question is: \textit{will the symbolic policy become unreliable if the OD module fails}? Therefore, in this section, we test out various types of suboptimal object detectors and measure the drop in performance. 
    
\begin{figure}
            \centering
            \includegraphics[width=0.45\textwidth]{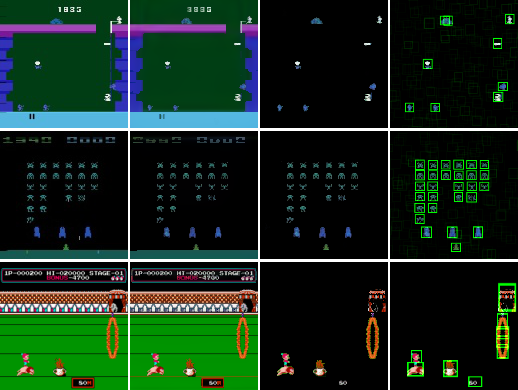}
            \caption{Visualizations of the object detection submodule in DiffSES, which is trained in an unsupervised way using SPACE \cite{lin2020space}. From left to right: two consecutive frames, detected foreground, the bounding boxes.}
            \label{fig:SPACE-collage}
        \end{figure}

    \textbf{Object Detection Visualizations.}
        We provide the object detection results in \Cref{fig:SPACE-collage}. It can be observed that even in these diverse scenarios with a combination of dynamic objects, the OD submodule trained without supervision is capable of detecting objects well.
        
    \textbf{Handling Under-Fitted Object Detection Module.}
    As part of our first case study, we emulate a poorly performing object detector by using intermediate training checkpoints of a fully trained object detector and compare its accuracy and rewards with its fully trained counterparts. Specifically, in the AdventureIsland3-Nes environment, we sample checkpoints of SPACE at 30\%, 50\%, and 80\% of its training. We then retrain the DiffSES on each of their outputs and compare their performances with a 100\% trained OD agent on the same. In \Cref{tab:od-failure}, we calculate the Average Precision for an IoU threshold of $50\%$ with respect to a ground-truth. The ground-truth is generated through a template-matching-based approach, which works robustly in gym/arari style games. Observing the results, we see that the rewards drops almost linearly with under-fitted OD module, i.e., when the OD submodule is only 80\% trained, the symbolic policy still achieves 82.4\% reward, and did not lead to an entire collapse of the algorithm.

        \begin{table}[h]
            \centering
            \begin{tabular}{c|c|c}
                \toprule[1.5pt]
                Intermediate OD Checkpoint & OD Avg. Precision & Reward \\
                \midrule
                $30$\% trained & $41.66$ & $1091$ \\ 
                $50$\% trained & $68.21$ & $2271$ \\ 
                $80$\% trained & $77.94$ & $2681$ \\ 
                $100$\% trained & $85.45$ & $3250$ \\ 
                \bottomrule[1.5pt]
            \end{tabular}
            \caption{Checkpoints of OD taken from the middle of training in the AdventureIsland3-Nes environment.}
            \label{tab:od-failure}
        \end{table}

    \textbf{Handling Simulated Object Missing.}
    We test the robustness of the learned symbolic policy by manually dropping the detected objects. We take the CircusCharlie as a case study, and drop both the crucial objects (ones that the Charlie will mainly interact) and non-crucial objects. The results in \Cref{tab:fragility} shows that the algorithm fails only when the crucial objects are dropped with significant probability.

    \begin{table}[h]
      \begin{minipage}{0.95\linewidth}
        \centering
        
        \caption{Simulated object detection failures in the CircusCharlie environment.}
        \label{tab:fragility}
        \resizebox{1\textwidth}{!}{
            \begin{tabular}{c|c|c|c}
                \toprule[1.5pt]
                \multicolumn{3}{c|}{\textbf{Dropping Probability}} & \multirow{2}{*}{\textbf{Reward Obtained} ($\uparrow$)} \\
                \cmidrule{1-3}
                Fire Ring (crucial) & Fire Pot (crucial) & Money Bag (passive) & \\
                \midrule
                $0.0$ & $0.0$ & $0.0$ & $7690$ \\
                $0.0$ & $0.0$ & $1.0$ & $7690$ \\
                $0.3$ & $0.3$ & $1.0$ & $4200$ \\
                $0.5$ & $0.5$ & $1.0$ & $3500$ \\
                $0.9$ & $0.9$ & $1.0$ &  $900$ \\
                \bottomrule[1.5pt]
            \end{tabular}
        }
        
      \end{minipage}
    \end{table}

\begin{figure}
            \centering
            \begin{subfigure}[b]{0.25\textwidth}
                \includegraphics[width=\textwidth]{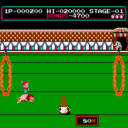}
                \caption{Input Image (Circus-Charlie)}
            \end{subfigure}
            \hspace{5ex}
            \begin{subfigure}[b]{0.25\textwidth}
                \includegraphics[width=\textwidth]{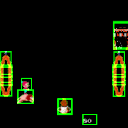}
                \caption{Segmented Foreground Objects}
            \end{subfigure}
            \caption{Erroneous Object Splitting in the case of ``Fire-Ring'' and ``Charlie'' in CircusCharlie-Nes. Observe that the ring gets broken up into four pieces and Charlie and his Lion get separated into two entities.}
            \label{fig:erroneous-object-splitting}
        \vspace{-2em}
        \end{figure}

    \textbf{Erroneous Object Splitting.}
        In a few instances of object detection, the SPACE algorithm tends to split a single object into multiple smaller parts as seen in \Cref{fig:erroneous-object-splitting}. It could be a flaw in the patch-based processing that SPACE employs. Although it can be prevented by careful hyper-parameter optimization of the boundary loss coefficients and bounding box sizes of the object detector, we do not attempt to overfit the hyperparameters towards any single environment in our setting, so as to maintain the notion of a truly end-to-end generalized framework with \textit{minimum handcrafting required}. 
        
        Despite such erroneous detection, DiffSES is in fact capable of learning robust rules for these environments as seen in \Cref{tab:additional-visual-rl}. This is because when one object is splitted, the policy actually learns to take only one fixed piece of the splitted parts. Therefore, the object splitting did not cause significant performance degredation.
        
    Overall ablation study results show that acceptable performance can be attained by DiffSES during certain failure cases of object detection. The object representations and the learned rules display reliability as a whole, even when the object detector is unable to retain its accuracy, the entire pipeline is capable of working with those broken or missing outputs and achieving satisfactory rewards.

\subsubsection{Ablation On The Fine Tuning Techniques}
\label{sec:exp:ablation:guided-search}

In the fine-tuning stage III, we developed two techniques: the \textit{Neural Guided Search} and the \textit{Differentiable Parameter Tuning}. In this section, we verify the acceleration as well as the performance improvement brought about by these techniques. 

\textbf{Removing The Neural Guided Search.} The ablation results for the Neural Guided (NG) search are provided in \cref{tab:ablation:guided-search}, where we train the two scenarios -- with and without neural guidance for DiffSES. For the one without neural guided search, the symbolic expressions mutate simultaneously for all $N_A$ trees in the forest without, similar to the CGP method. We compare these models based on their reward obtained, time taken (in seconds), and the number of generations required for convergence. Note that the model converges much slower without neural guided search as the total generations for training each $N_A$ tree increased.

\begin{table}[h]
    \centering
    \resizebox{0.52\textwidth}{!}{
    \begin{tabular}{c|c|c|c}
        \toprule[1.5pt]
        Configuration & Reward & Time (secs) & Generations \\
        \midrule
        Without Neural Guidance & $6.4$ & $7281$ & $103$ \\
        With NG (proposed) & $20.2$ & $5807$ & $82$ \\
        \bottomrule[1.5pt]
    \end{tabular}
    }
    \caption{Ablations based on Neural Guidance.}
    \label{tab:ablation:guided-search}
    \vspace{-1em}
  
    \end{table}

\textbf{Removing The Differentiability.} The ablation results for differentiability is shown in \Cref{tab:ablation:differentiability}. To perform this ablation study, we disable the SGD strategy during the symbolic expression evolution. From implementation side, we do not set the nodes in the symbolic tree to be trainable. We use genetic programming to evolve the expressions, and set the fitness metric as to maximize the obtained environment reward. 




As seen in \Cref{tab:ablation:differentiability}, the vanilla GP underperforms the proposed differentiable method, and also requires longer time to converge. The reason might be that when different trees' values are close, the decision threshold is easy to cross, hence subtle differences in the continuous values of the symbolic tree nodes can lead to different discrete decisions.

\begin{table}[h]
    \centering
    \resizebox{0.45\textwidth}{!}{
    \begin{tabular}{c|c|c|c}
        \toprule[1.5pt]
        Configuration & Reward & Time (secs) & Generations \\
        \midrule
        Vanilla GP & $12.2$ & $6275$ & $98$ \\
        GP+SGD (proposed) & $20.2$ & $5807$ & $82$ \\
        \bottomrule[1.5pt]
    \end{tabular}
    }
    \vspace{0.5em}
    \caption{Ablations based on Differentiable Training.}
    \label{tab:ablation:differentiability}
\vspace{-2em}    
\end{table}

\begin{table*}[t]
    \def\yes{\ding{52}}
    \def\no{\ding{56}}
    \def\na{\Flatsteel}
    \centering
    \resizebox{1\textwidth}{!}{
    \begin{tabular}{c|c|c|p{2cm}|c|c|p{6cm}}
    
    \toprule[1.5pt]

    \textbf{Methods} & \textbf{Visual} & \textbf{Non-Visual} & \textbf{Output Dims} & \textbf{Algorithm} &  \textbf{Differentiability}  & \textbf{Primitives (Expert Knowledge) Required} \\
    \midrule

    \multirow{2}{*}{DiffSES}  & \multirow{2}{*}{\yes} & \multirow{2}{*}{\yes} & \multirow{2}{*}{$N$-dim space} & \multirow{2}{*}{OD+GP+SGD}  & \multirow{2}{*}{\yes} & Basic object features (type, location, speed) and basic math operators ($+$, $-$, $\times$, $/$, $\geq$, $\neg$)  \\ 

    \cmidrule(r){2-7}
     \multirow{2}{*}{SDRL\cite{lyu2019sdrl}} & \multirow{2}{*}{\yes} & \multirow{2}{*}{\no} & $N$-dim continuous space & \multirow{2}{*}{Intrinsic Reward + Q Learning}  & \multirow{2}{*}{Specified Planning}  & High level representations (ladder, platform, rope, key, door, to open the door, etc.) \\

    \cmidrule(r){2-7}
    \multirow{4}{*}{Atari CGP \cite{wilson2018evolving}}  & \multirow{4}{*}{\yes} & \multirow{4}{*}{\no} & \multirow{4}{*}{$N$-dim space} & \multirow{4}{*}{CGP} & \multirow{4}{*}{\no} & Numerous specified math operators ($|x|^{p_n+1}$, $\frac{e^x-1}{e-1}$, $\mathrm{skewness}(x)$, $\mathrm{kurtosis}(x)$, ...), matrix operations ($\mathrm{first-element}(x)$, $\mathrm{split}(x)$, ...)   \\

    \cmidrule(r){2-7}
     NSRL\cite{ma2021learning} & \yes & \no & N-dim space & Transformer + ILP & \yes & Basic Object features \\
     
    \cmidrule(r){2-7}
    \multirow{2}{*}{Rollout IW \cite{bandres2018planning}}  & \multirow{2}{*}{\yes} & \multirow{2}{*}{\no} & $N$-dim continuous space & \multirow{2}{*}{Width based search } & \multirow{2}{*}{\no} & \multirow{2}{*}{ B-PROT features of screen pixels\cite{liang2015state}}\\
    
    \cmidrule(r){2-7}
    \multirow{1}{*}{VAE-IW \cite{dittadi2020planning}}  & \multirow{1}{*}{\yes} & \multirow{1}{*}{\no} & \multirow{1}{*}{N-dim space} & \multirow{1}{*}{variational autoencoders,Width based search} & \multirow{1}{*}{\no} & \multirow{1}{*}{B-PROT features of screen pixels\cite{liang2015state}}\\

    \cmidrule(r){2-7}
    \multirow{2}{*}{DSP \cite{landajuela2021discovering}}  & \multirow{2}{*}{\no} & \multirow{2}{*}{\yes} & $N$-dim continuous space & \multirow{2}{*}{RNN + Policy Gradient} & \multirow{2}{*}{Indirect} & Basic math operators ($+$, $-$, $\times$, $/$, $\mathrm{sin}$, $\mathrm{cos}$, $\log$, ... ),Basic Object feature, Pretrained neural network  \\

    \cmidrule(r){2-7}
    \multirow{1}{*}{PIRL \cite{verma2018programmatically}}  & \multirow{1}{*}{\no} & \multirow{1}{*}{\yes} & $N$-dim space & \multirow{1}{*}{Bayesian optimization} & \multirow{1}{*}{\yes} & 
    Basic object features, Pretrained neural network
    \\

    \cmidrule(r){2-7}
    \multirow{2}{*}{PLANQ-learning \cite{grounds2005combining}}  & \multirow{2}{*}{\no} & \multirow{2}{*}{\yes} & $N$-dim continuous space & \multirow{2}{*}{Q-learning, STRIPS} & \multirow{2}{*}{\na} & \multirow{2}{*}{\na}\\
    \bottomrule[1.5pt]
    \end{tabular}
    }

    \caption{Comparison across different symbolic RL methods.}
    \label{tab:baselines}
    \vspace{-2em}

    \end{table*}

\subsection{Comparison With Existing Symbolic RL Methods}
\label{sec:exp:baseline}

We consider the following three aspects as the most important properties of a symbolic RL approach: \ding{202} the \textit{simplicity} of the learned symbolic policy, \ding{203} the \textit{applicable domains} of the approach, and \ding{204} the \textit{amount of required human expert knowledge} to run the approach. The simplicity is vital as it is the most prominent advantage of symbolic RL over (neural network based) DRL: DRL is generally considered to be more flexible and has better potential to learn competitive policies than symbolic RL \cite{landajuela2021discovering}. Therefore, the merit of a symbolic RL approach will be deprecated if it could not be simpler, easier to understand and more efficient to execute than the DRL. On the other hand, the applicable domains of the approach and the amount of required expert knowledge directly dictate how well could this approach scale into more diverse and complex RL domains. A thorough comparison of these properties among different symbolic RL approaches as well as the proposed DiffSES is displayed in \Cref{tab:baselines}. 

\vspace{-1em}
\subsubsection{Non-visual RL Settings}

First, we compare our proposed method with Deep Symbolic Policy (DSP) \cite{landajuela2021discovering} on three continuous control-based environments -- CartPole, MountainCar, and Pendulum. The DSP method employs a risk-seeking policy gradient to maximize the performance of the generated policies. In this way, it is similar to DiffSES, which uses the PPO algorithm and genetic programming to optimize a symbolic tree. The difference is that for DSP the learning happens at an sequence predictor, where for DiffSES it happens directly at the expressions, as discussed in \Cref{sec:related works}.

Since DSP is incompatible with discrete output spaces, we use CartPoleContinuous, a continuous action space version of the original CartPole. We also include conventional DRL methods such as PPO and A2C to compare performance. As the intrinsic behavior of control-based continuous environments is rule-based, symbolic methods have no trouble converging on performant solutions. 

Results are provided in \Cref{tab:non-visual-comparison}. The resulting policies of DiffSES and DSP are both white-box and beats conventional approaches such as PPO and A2C, with DiffSES holding the upper hand.

\begin{figure*}[t]
    \centering



    \begin{minipage}[t]{0.9\textwidth}

    \centering
    \resizebox{1\textwidth}{!}{

    \begin{tabular}{c?c|c?c|c|c|c}
        \toprule[1.5pt]
        \multirow{2}{*}{\textbf{Environment}} & \multicolumn{2}{c?}{\textbf{Resultant Policy}} & \multicolumn{4}{c}{\textbf{Reward obtained}} \\
        \cmidrule{2-3}
        \cmidrule{4-7}
         & DSP & DiffSES & A2C & PPO & DSP & DiffSES \\
        \cmidrule{1-7}
        CartPole & $10.03 s_3 + 0.45 s_4$ & $s_2 + 2 s_3 + 3 s_4$ & $416$ & $6737$ & $8442$ & $10000$ \\
        MountainCar & $0.02 - \frac{0.72}{log(s_2)}$ & $\frac{s_2}{0.175}$ & $96.28$ & $94.71$ & $98.55$ & $99.33$ \\
        Pendulum & $-7.08 s_2 - \frac{13.39 s_2+3.12 s_3}{s_1} + 0.27$ & $\frac{(3 s_2 + 0.618 s_3) s_1}{-0.107}$ & $-165.99$ & $-153.85$ & $-151.21$ & $-119.05$ \\
        \bottomrule[1.5pt]
    \end{tabular}


    }
    \captionof{table}{Comparison of DiffSES with DSP \cite{landajuela2021discovering}, a non-visual Symbolic RL method.}
    \label{tab:non-visual-comparison}
    \end{minipage}
\end{figure*}

\vspace{-1em}

\subsubsection{Visual RL Settings}

Symbolic solutions in a Visual RL setting are the primary aim of our work. We compare the performance of DiffSES with Cartesian Genetic Programming (CGP) \cite{wilson2018evolving} on the Pong and SpaceInvaders environments, and the results are shown in \Cref{tab:visual-comparison}. 

\begin{table}[h]
    \centering
    \resizebox{0.5\textwidth}{!}{
    \begin{tabular}{c?c|c|c|c}
        \toprule[1.5pt]
        \multirow{2}{*}{\textbf{Environment}} & \multicolumn{4}{c}{\textbf{Reward obtained}} \\
        \cmidrule{2-5}
         & CGP & DiffSES & A2C & PPO \\
        \cmidrule{1-5}
        Pong & $19.7$ & $20.2$ & $17.2$ & $20.9$ \\
        SpaceInvaders & $713.60$ & $792.39$ & $627.1$ & $960.3$ \\
        \bottomrule[1.5pt]
    \end{tabular}
    }
    \caption{Comparison of DiffSES with CGP \cite{wilson2018evolving}, a Visual Symbolic RL method. }
    \label{tab:visual-comparison}
  \vspace{-2em}
  \end{table}

The DiffSES shows slightly better performance than the CGP in these environments. Besides the results, in terms of model simplicity, CGP applies matrix operators on pixel-level inputs, while the DiffSES uses the foreground object coordinates and velocities as operands, which may be easier to interpret.

\subsection{Comparison With Deep Network And Humans}
\label{sec:compare-with-drl-human}
While competing with deep neural networks based agent (DRL) is \textit{not} the intention of this work, in this section, we offer a brief comparison with the DRL, the teacher model of DiffSES, as well as human players. We found that the distilled symbolic policy can differ from the DRL on both the performance and the transferability.

\begin{figure*}[!h]
\vspace{-1em}
    \begin{center}
    \centerline{ \includegraphics[width=0.7\columnwidth]{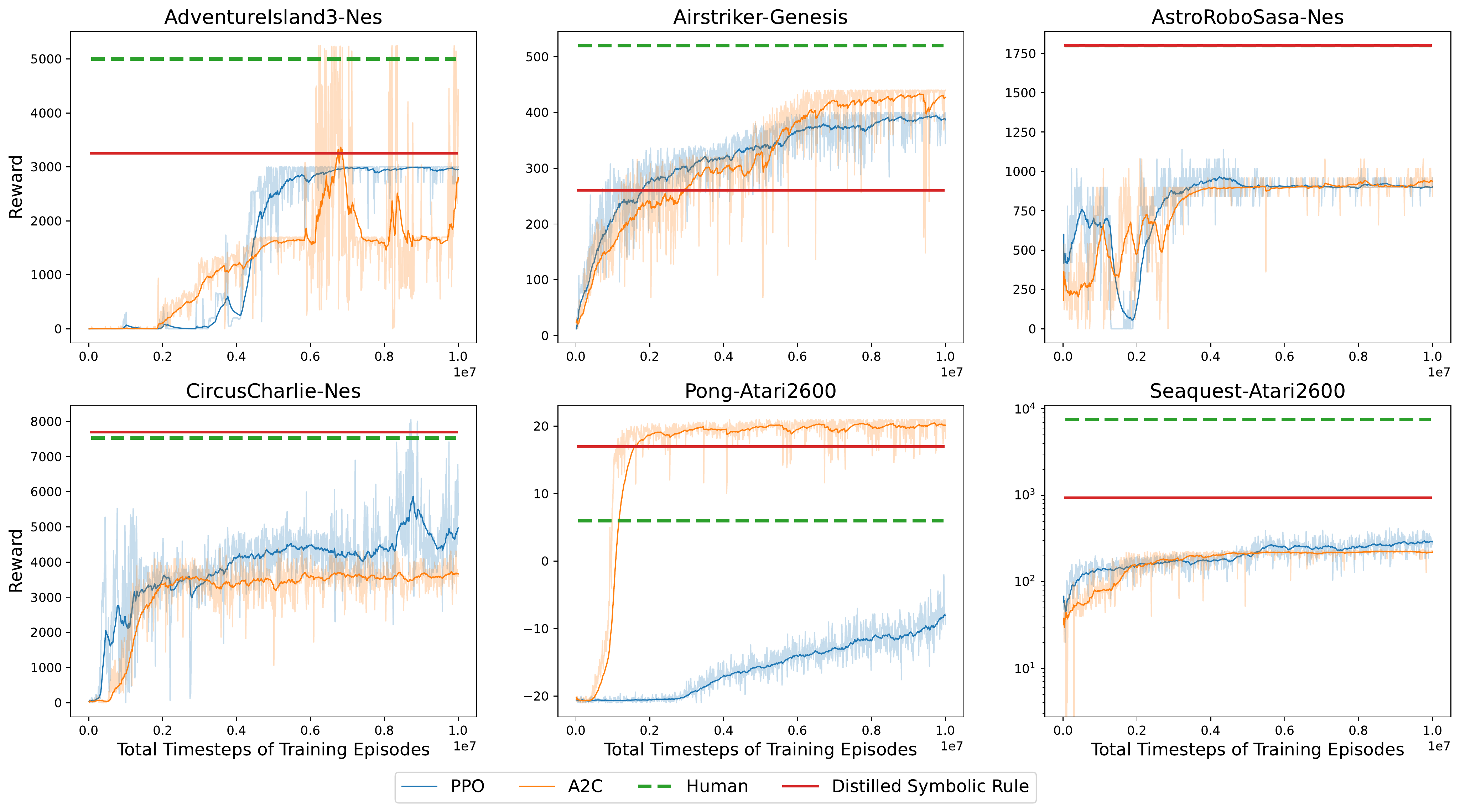}}
    \caption{Performance validation for the CNN-based RL agents and the learned (distilled) symbolic expression.}
    \label{fig:exps}
    \vspace{-2em}
    \end{center}
    \end{figure*}

\subsubsection{Performance Comparison}
The reward performances for DiffSES and its PPO teacher model are shown in \Cref{tab:additional-visual-rl}. Additionally, in \Cref{fig:exps}, we plotted the accumulated reward for PPO~\cite{ppo}, A2C~\cite{mnih2016asynchronous}, human player, and the symbolic policy. The human player scores are obtained by playing under a lowered frame rate. The symbolic policies in some cases perform better than the DRL. Especially, when there exists simple underlying rules (e.g. in the Pong environment, the racket should go to the aiming point of the pong), DiffSES could learn these rules in a concise way, while the DRL struggles to approximate them with heavy coefficients.



\subsubsection{Behavior Comparison}
\begin{figure*}[h]
    \vspace{-1em}
    \centering
        \includegraphics[width=0.3\textwidth]{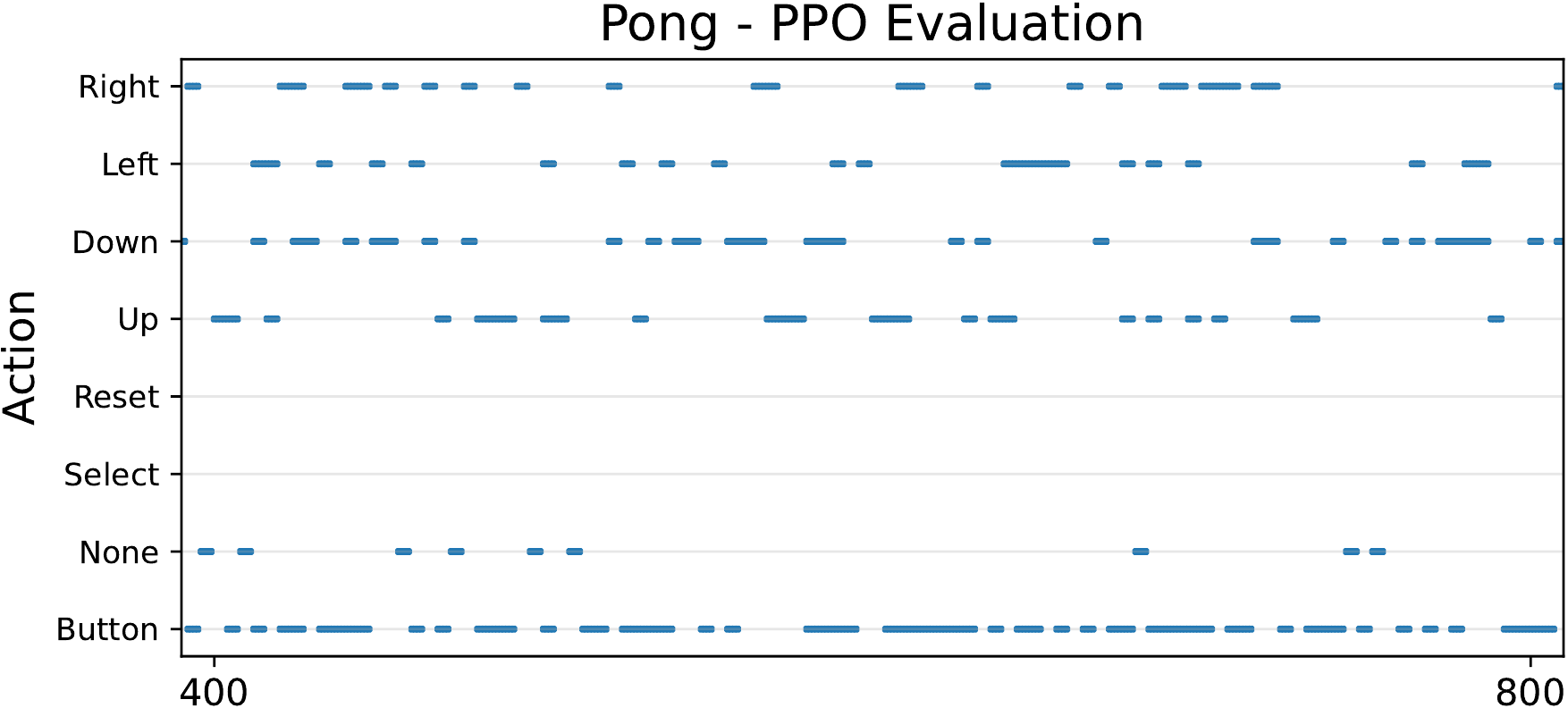}
        \includegraphics[width=0.3\textwidth]{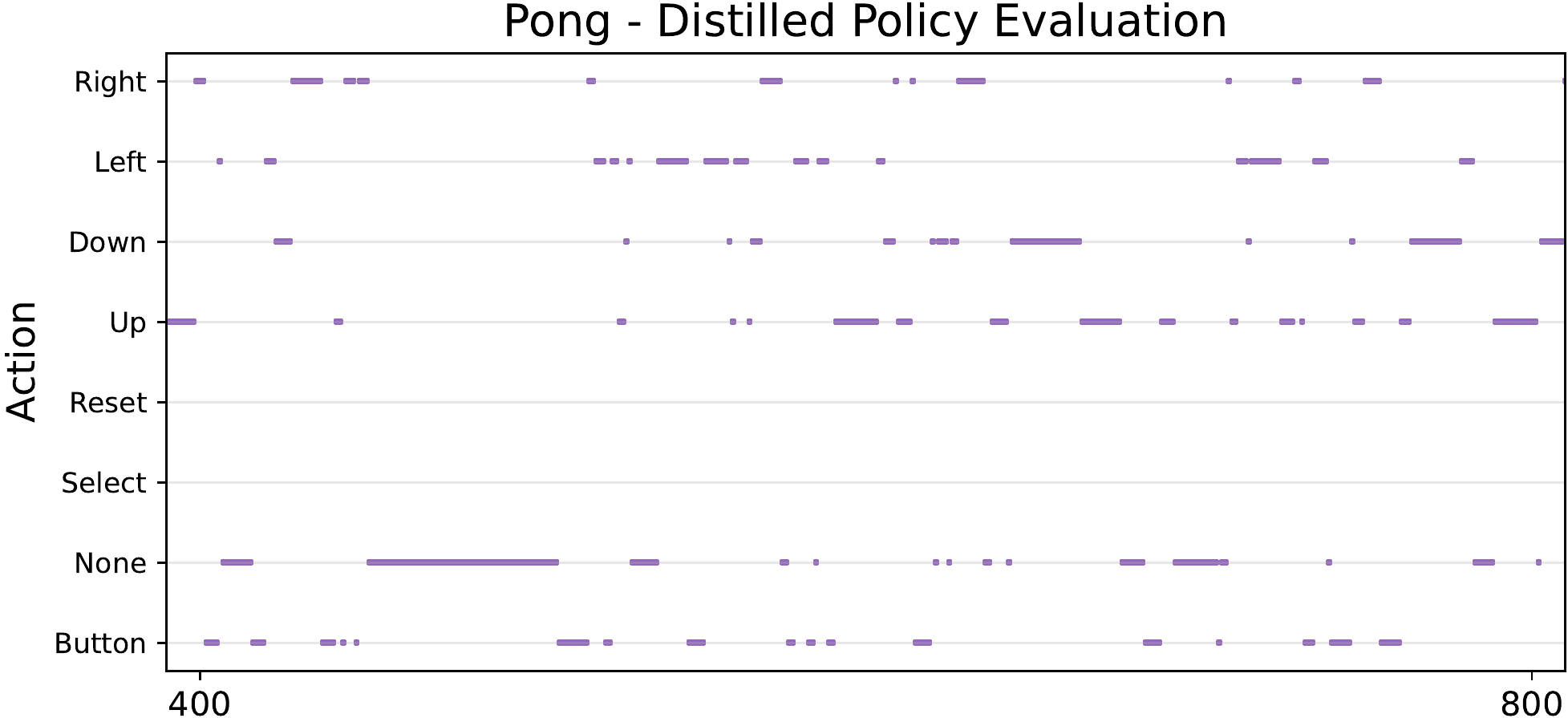}
    \caption{Action distributions of the PPO teacher agent (left) and the learned symbolic policy distilled and learned out of the PPO teacher (right).}
    \label{fig:behaviors}
    \vspace{-1em}
    \end{figure*}

We show the behaviors of the original DRL, as well as the symbolic policies under these two dataset generation schemes in \Cref{fig:behaviors}. We found that the learned symbolic policy displays different trajectory than the teacher DRL model, for the same initial state.

\subsubsection{Policy Transfer Comparison}

\begin{figure}
        \begin{minipage}[t]{1\textwidth}
            \centering
            \begin{subfigure}[h]{0.3\textwidth}
                \includegraphics[width=\textwidth]{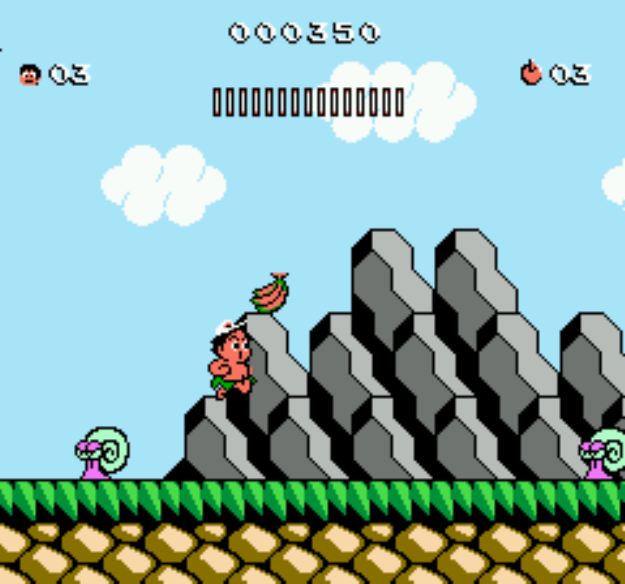}
                \caption{AdventureIsland3}
            \end{subfigure}
            \hspace{5ex}
            \begin{subfigure}[h]{0.3\textwidth}
                \includegraphics[width=\textwidth]{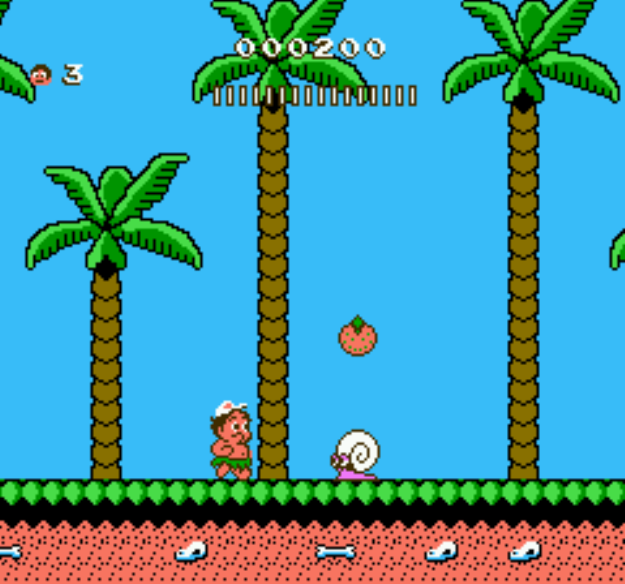}
                \caption{AdventureIsland2}
            \end{subfigure}
            \captionof{figure}{Testing on the policy transferrability. The two images are screenshots of the two tested environments that have similar but not identical styles. The PPO and A2C agents are trained on AdventureIsland3 (AI-3), tested on AdventureIsland2. The symbolic policy is learned based on the PPO agent trained on AdventureIsland3. Neither PPO/A2C nor symbolic policies/OD submodule are fine-tuned/modified on AdventureIsland2.}
            \label{fig:od-transfer}
        \end{minipage}

        \begin{minipage}[t]{1\textwidth}

            \centering
            \resizebox{1\textwidth}{!}{
            \begin{tabular}{c|c | c  | c  | c|c|c|cc c   }
            \toprule[1.5pt]
            \textbf{Evaluation on} & \multicolumn{1}{c|}{\textbf{\makecell{ AdventureIsland3}}} & \multicolumn{6}{c}{\textbf{\makecell{AdventureIsland2}}} \\ 
            \cmidrule(r){1-1} \cmidrule(r){2-2} \cmidrule(r){3-8} 
            Checkpoints from AI-3 & 10M & 1.0M & 2.5M & 5.0M &  7.5M & 10M & Best \\
               \cmidrule(r){1-8} 
            A2C  & \textbf{5150} & 0  & 200 & 50 & 100 & 100 & 200 \\
            PPO (teacher DRL) & 3000 & 150  & 0 & 100 & 50 & 50 & 50 \\
            Learned Symbolic Policy & 3250 (OD presci. 85.45) & \multicolumn{6}{c}{\textbf{1950} (OD presci. 52.21)} \\
            \bottomrule[1.5pt]
            \end{tabular}
            }
            \caption{The results of the transferrabilty case study.}
            \label{tab:transfer}

        \end{minipage}

        \end{figure}

We tested the transferability of different policies under domain shift. We run experiments in the AdventureIsland2 and AdventureIsland3 environments (Refer \Cref{fig:od-transfer}). We first train a teacher PPO on AdventureIsland3 and learn the symbolic. We then directly test both the PPO and the symbolic expression on AdventureIsland2, without tuning the PPO coefficients nor changing the symbolic expression and the OD module. As can be seen in Table~\ref{tab:transfer}, the CNN models completely fail to match their performances in the original environment, over the checkpoint from the entire training history. On the contrary, though the OD module performance dropped in the new environments, the symbolic policy still transfers better, with significantly fewer performance loss. 

The performance gain of symbolic policy may come from the disentanglement of object detection and action inference: if in two different scenarios, the scenes' pixel-level attributes are markedly different while the logical specifications are congruent, the OD will naturally group them together. On the other hand, though both the DRL and OD can fail in new visual scenes, the DRL policy fails as a whole. Therefore, if we isolate the OD from the symbolic expression, the symbolic expression transfers better than the end-to-end DRL.

\subsection{More Examples of learned equations}
\label{sec:appendix}
\begin{figure*}
    \centering
    \includegraphics[width=0.7\textwidth]{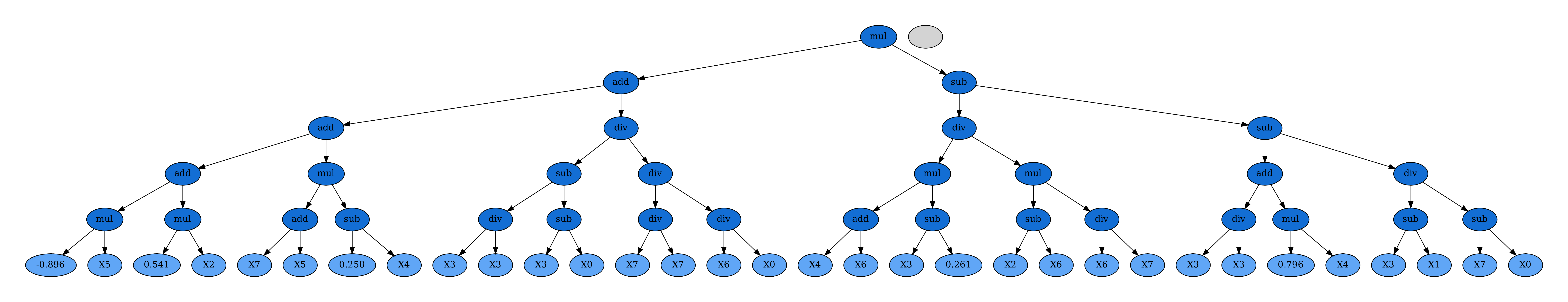}
    \caption{Visualization of a trained DiffSES expression tree for the CircusCharlie-Nes environment}
    \label{fig:policy-circuscharlie}
\end{figure*}

\begin{figure*}
    \centering
    \includegraphics[width=0.9\textwidth]{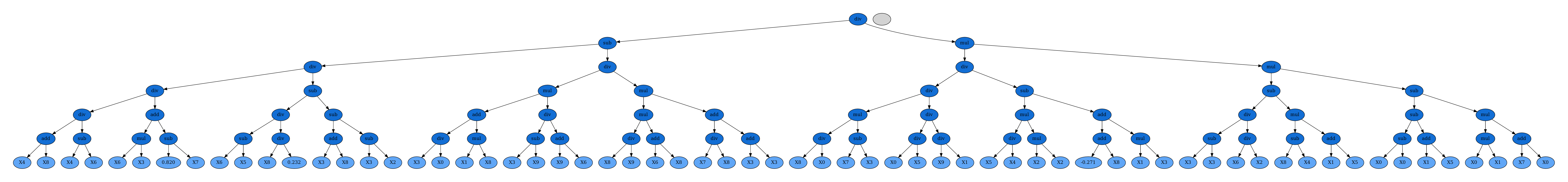}
    \caption{Visualization of a trained DiffSES expression tree for the Seaquest-Atari2600 environment}
    \label{fig:policy-seaquest}
\end{figure*}

More examples of the learned symbolic controller expressions are displayed in \Cref{fig:policy-seaquest} and \Cref{fig:policy-circuscharlie}. We observe a tradeoff between symbolic expression simplicity and agent ability/environment complexity, as the the generated symbolic trees for CircusCharlie-Nes and the Seaquest-Atari2600 are more complex, and requesting more object information to perform effective control.

\section{Discussions And Limitations}
\label{sec:limitations}
This work aims to improve current symbolic reinforcement learning (RL) methods by reducing human expert knowledge and making the policy simpler more scalable to complex visual scenes. To the best of our knowledge, this is the first work to perform a differentiable symbolic search for a visual RL domain and the first to base operator level symbolic policies on object representations.

The proposed symbolic expressions perform better than previous symbolic RL methods, but this comes with the assumptions of having an expert policy available for distillation and having a neural network-based object detector available. While the proposed approach scales well to Atari/Retro/Gym-style image inputs with the help of partially differentiable optimization, its broader applicability to more complex scenarios such as real-world 3D vision inputs has yet to be optimized and verified. In future work, we plan to improve the neural-symbolic co-evolution and test the approach in more difficult scenarios.

We also find that the learned symbolic policies tend to become less interpretable as the environment becomes harder and/or the model performance improves. We note this as a possible tradeoff between expression simplicity and model performance.

\section{Conclusion}

This paper proposes Differentiable Symbolic Expression Search (DiffSES), a novel symbolic reinforcement learning framework that generates simple and competitive symbolic policies composed of symbolic operators and object representations. Compared with previous symbolic RL methods, the proposed approach requires a smaller and simpler math operatoer set, hence significantly reduces the need for human expert knowledge in the design process and scales better to complex, high-dimensional visual inputs. Additionally, the proposed neural-guided search augments the symbolic policy evolution process, by introducing flexible optimization and differentiability to the existing genetic programming algorithm. Our approach paves the way towards learning more flexible symbolic policies in complex reinforcement learning domains.

\vskip 0.2in

\bibliography{refs}

\end{document}